\documentclass[]{ceurart}
\usepackage[utf8]{inputenc}
\usepackage{csquotes}
\usepackage{booktabs}
\usepackage{amsmath}
\usepackage{amssymb}
\usepackage{graphicx}
\usepackage{caption}
\usepackage{float}
\usepackage{microtype}
\usepackage{tikz}
\usetikzlibrary{shapes.geometric,arrows.meta,positioning}

\begin{document}

\title{IGT @ FinMMEval 2026 Task 2: Question-Type Prompting\\
       with Targeted Extraction for Multilingual Financial QA}

\author[1]{Yuwen Chiu}[%
  orcid=0009-0008-2859-4679,
  email=ychiu60@gatech.edu
]
\cormark[1]
\address[1]{Georgia Institute of Technology, North Ave NW, Atlanta, GA 30332}
\cortext[1]{Corresponding author.}

\copyrightclause{Copyright for this paper by its authors.
Use permitted under Creative Commons License Attribution 4.0
International (CC BY 4.0).}

\begin{abstract}
We present the IGT system for PolyFiQA Task~2 of the FinMMEval
Lab at CLEF~2026, a multilingual financial question answering
task over English SEC filings and multilingual news articles
(English, Chinese, Japanese, Spanish, Greek) for four companies.
Our central observation is that the $344$ development questions
divide into two families requiring fundamentally different
approaches: structured numeric types (R\&D ratio, cash flow,
capital expenditure) are best answered by direct keyword
extraction on filing text, while synthesis types (investment
strategy, capital allocation, top-three revenue focuses) require
rule-based multilingual news passage selection. A dataset analysis
reveals that $17$--$18$ of $19$ ground-truth reference answers
per synthesis type share an exact evidence label prefix, whose
unigram tokens contribute directly to ROUGE-1 overlap. The final
system achieves development ROUGE-1~${\approx}\,0.395$, a $60\%$
relative improvement over a generic RAG baseline
(${{\sim}}0.247$), and ranks 3rd of $12$ teams on the official
test set with ROUGE-1~$= 0.3071$, Precision~$= 0.2821$, and
Recall~$= 0.4044$.
\end{abstract}


\conference{CLEF 2026 Working Notes, 21--24 September 2026, Jena, Germany}

\maketitle

\section{Introduction}

Financial question answering presents a deceptively heterogeneous
problem. Even within a single benchmark, questions asking ``what were
the cash flows?'' and ``what is the company's investment strategy?''
require fundamentally different computational approaches: the first
has a deterministic answer derivable by pattern matching on a table;
the second requires synthesizing signals across multilingual news
sources with no single correct phrasing. Systems that treat these
uniformly (as most retrieval-augmented generation (RAG) pipelines do) pay a performance cost on both.

PolyFiQA Task~2, introduced at FinMMEval CLEF~2026~\cite{FinMMEval2026,FinMMEvalTask2Overview2026}, makes
this heterogeneity explicit. The task pairs English SEC filings with
multilingual news articles (English, Chinese, Japanese, Spanish, Greek)
for four companies and asks systems to produce concise evidence-grounded
answers evaluated by ROUGE-1~\cite{lin2004}. Our analysis of the
development set reveals eight distinct question types with different
primary evidence sources, answer structures, and even different
ground-truth prefix conventions that affect ROUGE-1 scores directly.

We begin from a diagnostic standpoint: we first study what the data
requires, then build accordingly. An initial exploratory analysis of
the development set motivated a KG-RAG pipeline with a learned
retrieval policy, which revealed that generation quality and output
format control dominated retrieval sophistication on this task. That
finding drove the final system design: routing questions to specialized
handlers, computing structured financial ratios directly rather than
retrieving them, and preserving non-English source text rather than
translating it. Each decision is motivated by a specific observed
failure mode.

The contributions of this work are:
\begin{itemize}
  \item Question-type routing to eight tailored handlers, with
        keyword extraction on raw filing text for structured numeric
        types and rule-based multilingual news passage selection for
        synthesis types.
  \item Company-specific label normalization for cash flow, R\&D, and
        capex extraction, and original-language revenue quote
        preservation for multilingual evidence.
  \item A systematic ablation across all design decisions with
        per-type and per-company performance analysis, pseudocode
        for key extraction routines, and full prompt templates
        in Appendix~\ref{app:prompts}.
\end{itemize}

\section{Related Work}

Financial document QA has received sustained attention, motivated by
the scale and complexity of regulatory filings. The most studied setting
is English-only retrieval over 10-K and earnings call transcripts, where
RAG pipelines face two recurring challenges: chunking strategies that
destroy table structure, and context windows too short to hold a full
filing. Guo et al.\ showed that element-based chunking (preserving
table and title boundaries) outperforms fixed-size splits, and that
cross-encoder re-ranking and HyDE query expansion~\cite{gao2022hyde} provide further gains,
though all zero-shot methods fall well short of an oracle that always
retrieves the correct passage~\cite{guo2024improving}. Kim et al.\
addressed scale in the ACM-ICAIF 2024 FinanceRAG challenge with a
two-stage multi-reranker pipeline and documented a ``32k token wall''
beyond which LLM generation quality degrades sharply~\cite{kim2024multireranker}.

The multilingual dimension of PolyFiQA is largely unexplored in prior
financial QA work. Lef{\'e}bure et al.\ studied Spanish--English
bilingual financial LLMs and found that instruction tuning on Spanish
data unexpectedly improves English performance through cross-linguistic
transfer~\cite{lefebure2024dolares}, but their setting translates
everything to a common language rather than preserving source-language
text. Xu et al.\ introduced FinRAGBench-V, a bilingual Chinese--English
benchmark for multimodal financial RAG, finding that multimodal
retrievers substantially outperform text-only baselines on documents
heavy in charts and tables~\cite{xu2024finragbench}. Our task differs
from both: evidence is multilingual across five scripts, but answers
are generated in English, requiring cross-lingual evidence grounding
without translation.

On the evaluation side, Mirza et al.\ provide the most relevant
diagnostic for our design choices~\cite{mirza2024lclm}. They show
that GPT-4o and GPT-4-Turbo are highly sensitive to prompt placement
and formatting on financial retrieval tasks, that performance collapses
at context lengths beyond 32k tokens for multi-concept questions, and
that standard Recall metrics artificially inflate apparent performance
compared to F1. These findings directly inform our use of format-specific
prompts and per-type token budgets.

For multi-hop reasoning, Asai et al.\ demonstrated that graph-based
recurrent retrieval over Wikipedia hyperlinks substantially outperforms
single-step retrieval on multi-hop QA benchmarks~\cite{asai2020graph}.
We test an analogous two-hop approach on PolyFiQA Expert questions in
Section~5 and find it does not transfer: format-controlled direct
prompting outperforms multi-hop HyDE retrieval by $0.087$ ROUGE-1 on
the questions where multi-hop was designed to help.

\section{Task and Data Analysis}

PolyFiQA Task~2~\cite{peng2025multifinben} provides an English SEC filing excerpt $R$ and a set
of multilingual news articles $N = \{N_{\text{en}}, N_{\text{zh}},
N_{\text{ja}}, N_{\text{es}}, N_{\text{el}}\}$ for the same company.
Given question $Q$, the system generates answer $A$ of at most 100~words.
The dataset covers Microsoft (MSFT), Honeywell (HON), Johnson \& Johnson
(JNJ), and Universal Corporation (UVV), with 344~instances split evenly
between PolyFiQA-Easy (172 factual questions) and PolyFiQA-Expert
(172 analytical questions). The primary metric is ROUGE-1
F-measure~\cite{lin2004}. We refer readers to the lab overview~\cite{FinMMEval2026}
and task overview~\cite{FinMMEvalTask2Overview2026} for full task details.
Iterative development and all ablation experiments reported in
Section~5 use a fixed 152-question subset of these 344 development
instances, selected to give balanced coverage of all eight question
types across all four companies; the remaining development instances
were held out from tuning to limit overfitting to development-set
phrasing.

Before building any system, we analyzed the development set to understand
what each question actually requires. Two findings shaped our entire design.

\paragraph{Finding 1: Eight question types, two distinct information regimes.}
Clustering questions by phrasing and answer structure reveals eight
types that fall into two families. Structured numeric types (revenue
amount, cash flow, R\&D ratio, balance sheet, capital expenditure)
have answers derivable from a specific line item or computed ratio in
the SEC filing. There is one correct number, and finding it is a
retrieval precision problem. Synthesis types (investment strategy,
top-three revenue focuses, capital allocation) have answers requiring
multi-document reasoning across multilingual news, with no single
correct phrasing, and performance is bounded by how well the model
synthesizes and formats its response.

\paragraph{Finding 2: Ground-truth evidence labels are shared unigrams.}
Examining reference answer structure reveals that virtually every
instance begins with one of two evidence label prefixes:
\emph{News Evidence:} when the answer draws on news articles, or
\emph{Financial Statement Evidence:} when it draws on the filing.
Table~\ref{tab:evlabels} shows the distribution. These prefix tokens
are counted by ROUGE-1, meaning that generating the wrong label
eliminates up to three unigrams of overlap before any content is
compared. This observation motivates the FSE label assignment described
in Section~4.

\begin{table}[h]
\caption{Evidence label distribution by question type. NE~=~News
Evidence, FSE~=~Financial Statement Evidence. Each type contains
19 questions. FSE column includes one instance using the plural
variant \emph{Financial Statements Evidence:}.}
\label{tab:evlabels}
\begin{tabular}{lrr}
\toprule
Question Type & NE & FSE \\
\midrule
Revenue amount      & 18 &  0 \\
Cash flow           & 17 &  1 \\
R\&D ratio          & 17 &  1 \\
Balance sheet       & 18 &  0 \\
Investment strategy &  1 & 17+1 \\
Top-three focuses   &  0 & 18+1 \\
Capital expenditure &  1 & 17+1$^{\ddagger}$ \\
Capital allocation  &  0 & 18+1 \\
\bottomrule
\multicolumn{3}{p{7.5cm}}{\footnotesize $^{\ddagger}$Ground-truth
references for capital expenditure are predominantly FSE-labeled
($17{+}1$ of $19$), but the capex prompt instructs the model to begin
with \emph{Financial Statement Evidence:}, producing a systematic
label match with the majority of references. The single NE instance
represents a label mismatch that contributes to the capex ROUGE-1
floor (Section~5.2).}\\
\end{tabular}
\end{table}

\section{System Description}

\subsection{Overview}

Before arriving at the final architecture, we explored a substantially
more complex pipeline: a KG-RAG system using Weaviate vector retrieval
with a learned GRPO policy~\cite{shao2024grpo} that selected among six
retrieval strategies at inference time (semantic-only, entity-first,
document-type-first, pattern-first, balanced, and multi-document), with
Mistral-7B-Instruct-v0.2 as the generator. Bidirectional entity edges
in the knowledge graph increased retrieved path coverage from $68$ to
$103$ edges per instance, and multilingual query expansion provided
marginal additional gains. The best result from this pipeline was
ROUGE-1~${\approx}\,0.298$ on the development set. Switching to
Claude~Sonnet~4 with direct keyword extraction --- without any learned
retrieval component --- exceeded this by $+0.097$ ROUGE-1. The result
confirmed that generation quality and output format control dominated
retrieval sophistication on this task; all further development proceeded
with the simpler architecture described below.

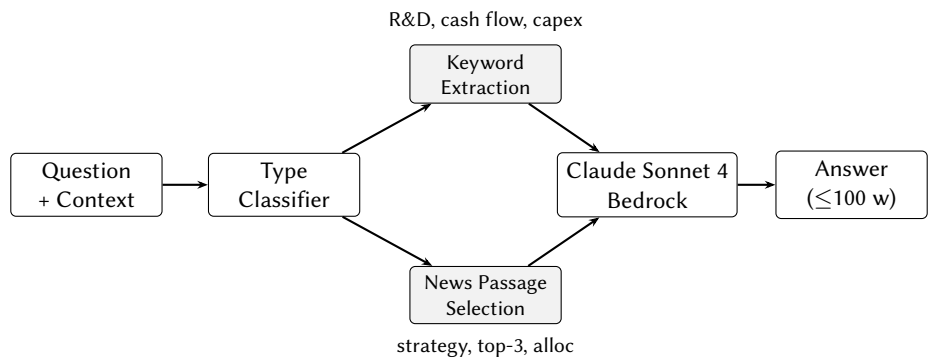
\begin{figure}[ht]
\centering
\begin{tikzpicture}[
  node distance=0.45cm and 0.6cm,
  box/.style={rectangle, draw, rounded corners=2pt,
              minimum width=2.0cm, minimum height=0.65cm,
              font=\footnotesize, align=center},
  solidbox/.style={rectangle, draw, rounded corners=2pt,
                   minimum width=2.0cm, minimum height=0.6cm,
                   font=\scriptsize, align=center, fill=gray!10},
  lbl/.style={font=\scriptsize},
  arr/.style={-{Stealth[length=4pt]}, thick}
]
\node[box] (q) {Question\\+ Context};
\node[box, right=0.6cm of q] (cls) {Type\\Classifier};
\node[solidbox, above right=0.65cm and 0.65cm of cls] (ext)
      {Keyword\\Extraction};
\node[lbl, above=0.05cm of ext] {\scriptsize R\&D, cash flow, capex};
\node[solidbox, below right=0.65cm and 0.65cm of cls] (sel)
      {News Passage\\Selection};
\node[lbl, below=0.05cm of sel] {\scriptsize strategy, top-3, alloc};
\node[box, right=2.6cm of cls] (llm) {Claude Sonnet~4\\Bedrock};
\node[box, right=0.5cm of llm] (out) {Answer\\(${\leq}$100 w)};
\draw[arr] (q) -- (cls);
\draw[arr] (cls) -- (ext);
\draw[arr] (cls) -- (sel);
\draw[arr] (ext) -- (llm);
\draw[arr] (sel) -- (llm);
\draw[arr] (llm) -- (out);
\end{tikzpicture}
\caption{Final system pipeline. The type classifier routes questions
to keyword extraction on raw filing text (structured numeric types)
or rule-based news passage selection (synthesis types), then
generates answers via Claude~Sonnet~4. Neither route uses vector
retrieval; the RAG baseline used in early development was replaced
after analysis showed dense retrieval fails on pipe-delimited
financial table chunks (Section~4, Targeted Extraction).}
\label{fig:pipeline}
\end{figure}

Figure~\ref{fig:pipeline} shows the system architecture. Every question
is classified by keyword matching into one of eight types and routed
to a type-specific handler. Structured numeric questions go to targeted
extraction functions that bypass retrieval entirely; synthesis questions
go to a context-assembly routine that selects relevant multilingual
passages and prepends the FSE label. Both paths call Claude~Sonnet~4~\cite{anthropic2025sonnet} via
AWS~Bedrock, with a maximum output of $300$ tokens per response.\footnote{Code
available at \url{https://github.com/chiuyuwen91/FinMMEval-2026}.}

\subsection{Question-Type Routing}

Table~\ref{tab:routing} summarizes the routing logic. The FSE label
is prepended for all synthesis types, aligning with Table~\ref{tab:evlabels}.
For top-three questions, the prompt specifies an exact format string
(\emph{``The top three revenue focuses from the news are: 1)''}) that
appears verbatim in ground-truth references, contributing additional
unigram overlap beyond the label tokens alone.

Routing is implemented as a priority-ordered keyword scan over the
question string.\footnote{Representative trigger phrases per type:
R\&D ratio --- \emph{``R\&D ratio,'' ``R\&D divided by''};
cash flow --- \emph{``cash flow,'' ``irregularities in the''};
revenue amount --- \emph{``revenue,'' ``total revenue,'' ``net revenue''};
balance sheet --- \emph{``balance sheet,'' ``total assets,'' ``total
liabilities''};
capital expenditure --- \emph{``capital expenditure,'' ``capex,''
``capital spending''};
investment strategy --- \emph{``investment strategy,'' ``strategic
priorities,'' ``strategy for''};
top-three focuses --- \emph{``top three,'' ``top-three''};
capital allocation --- \emph{``allocating capital,'' ``capital allocation''}.
The cash flow pattern is the most ambiguous: Expert questions
containing ``cash flow'' but requiring multi-document analytical
reasoning are occasionally mis-routed to the extraction path
(Section~5.3).}

\begin{table}[h]
\caption{Question-type routing and handling strategies.}
\label{tab:routing}
\begin{tabular}{p{3.2cm}p{2.2cm}p{4.4cm}}
\toprule
Question Type & Primary Source & Strategy \\
\midrule
Revenue amount      & Filings + news & Multilingual quote preservation \\
Cash flow           & Filings        & Keyword extraction + normalization \\
R\&D ratio          & Filings        & Direct ratio computation \\
Balance sheet       & Filings + news & Trend extraction \\
Capital expenditure & Filings + news & Capex regex + news context \\
Investment strategy & News           & FSE label; synthesis prompt \\
Top-three focuses   & News           & FSE label; exact format string \\
Capital allocation  & News           & FSE label; allocation prompt \\
\bottomrule
\end{tabular}
\end{table}

\subsection{Targeted Financial Extraction}

The most instructive failure of the initial RAG baseline was on cash
flow questions. Inspection of the top-5 retrieved chunks for a
Johnson \& Johnson cash flow question revealed that all five were
pipe-delimited table header rows from the SEC filing, where formatting
characters and whitespace dominate the vector representation rather
than financial content. The model received no dollar figures. This
failure is structural: fixed-size chunking splits financial statement
tables across chunk boundaries, and cosine similarity on a
multilingual sentence encoder reliably retrieves the section header
rather than the data rows beneath it. Better embeddings or re-ranking
cannot resolve a mismatch between the chunk format and the query intent.

We replaced retrieval with three direct extraction routines that
operate on the raw filing text.

\paragraph{Cash flow extraction.}
Extracts operating, investing, and financing cash flow totals using
company-specific keyword patterns. The need for company-specific
patterns became clear from JNJ's filings: JNJ uses non-standard
section headers that do not match the GAAP label patterns effective
for MSFT and HON, causing the generic extractor to retrieve figures
from the wrong cash flow category in approximately $30\%$ of JNJ
instances. Adding JNJ-specific keyword variants recovered ${\approx}0.038$ ROUGE-1 on JNJ cash flow questions.

\paragraph{R\&D ratio computation.}
Rather than asking the LLM to compute a ratio from retrieved text,
this function extracts R\&D expenditure and total revenue directly
and computes:
\begin{equation}
  \text{R\&D ratio} = \frac{\text{R\&D expenditure}}{\text{Total revenue}},
  \label{eq:rd}
\end{equation}
passing the result to the LLM as a scalar. The LLM's role is then
formatting rather than computation. This improved R\&D question
ROUGE-1 from ${\approx}0.28$ to ${\approx}0.582$, the largest
single per-type gain in the development process.

Appendix~\ref{app:pseudocode} gives the extraction logic in condensed
pseudocode form.

\paragraph{Capital expenditure extraction.}
A regex-based routine targeting capex line items, supplemented with
news passages filtered to 2{,}000~characters per language.

\subsection{Multilingual Revenue Quote Preservation}

A second instructive failure appeared on revenue amount questions.
Reference answers for these questions frequently quote revenue figures
in the original news source language (Japanese yen notation, Spanish billion phrasing) rather than translating to English. A system that
translates or paraphrases non-English passages before generation
produces answers that are factually equivalent but share no unigrams
with the reference. Our approach includes the most relevant news passage
per language verbatim in the prompt and instructs the model to preserve
source-language phrasing when quoting figures. This yields approximately
$+0.007$--$0.010$ ROUGE-1 on revenue questions and $+0.017$ overall.

\subsection{Token Budget Management}

\begin{figure}[ht]
\centering
\begin{tikzpicture}
\begin{scope}[xscale=0.9, yscale=14]
  \draw[->] (0,0.28) -- (6.6,0.28)
    node[right,font=\footnotesize]{News chars per language};
  \draw[->] (0,0.28) -- (0,0.68)
    node[above,font=\footnotesize]{ROUGE-1};
  \foreach \y/\lbl in {0.30/0.30,0.40/0.40,0.50/0.50,0.60/0.60}{
    \draw[gray!20] (0,\y) -- (6.4,\y);
    \node[left,font=\scriptsize] at (0,\y) {\lbl};
  }
  \foreach \x/\lbl in {
    0.5/800, 1.5/1000, 2.5/1200, 3.5/1500, 4.5/2000, 5.5/2500}{
    \node[below,font=\scriptsize] at (\x,0.28) {\lbl};
  }
  \draw[blue!70!black,thick]
    (0.5,0.5693)--(1.5,0.5369)--(2.5,0.5335)--(3.5,0.6173)--(4.5,0.5792);
  \foreach \x/\y in {0.5/0.5693,1.5/0.5369,2.5/0.5335,3.5/0.6173,4.5/0.5792}
    \draw[blue!70!black,fill=blue!70!black]
      (\x,\y) ellipse [x radius=0.09, y radius=0.006];
  \draw[green!55!black,thick,dashed]
    (0.5,0.3396)--(2.5,0.3391)--(3.5,0.3466)--(4.5,0.3569)--(5.5,0.3434);
  \foreach \x/\y in {0.5/0.3396,2.5/0.3391,3.5/0.3466,4.5/0.3569,5.5/0.3434}
    \draw[green!55!black,fill=green!55!black]
      (\x,\y) ellipse [x radius=0.09, y radius=0.006];
  \node[above,font=\tiny,blue!70!black] at (3.5,0.6173) {0.617$\star$};
  \node[above,font=\tiny,green!55!black] at (4.5,0.3569) {0.357$\star$};
  \draw[blue!70!black,fill=blue!70!black]
    (0.1,0.650) ellipse [x radius=0.09, y radius=0.006];
  \node[right,font=\scriptsize] at (0.3,0.650) {Revenue amount};
  \draw[green!55!black,thick,dashed] (2.8,0.650)--(3.2,0.650);
  \draw[green!55!black,fill=green!55!black]
    (3.0,0.650) ellipse [x radius=0.09, y radius=0.006];
  \node[right,font=\scriptsize] at (3.3,0.650) {Top-three focuses};
\end{scope}
\end{tikzpicture}
\caption{ROUGE-1 vs.\ news character budget per language for two
question types. Both curves are non-monotonic: revenue peaks at
$1{,}500$~chars and top-three at $2{,}000$~chars. Budgets beyond
the peak produce verbose outputs that are truncated before completing
key points, illustrating the coverage--verbosity trade-off that
motivates type-specific budgets.}
\label{fig:budget}
\end{figure}
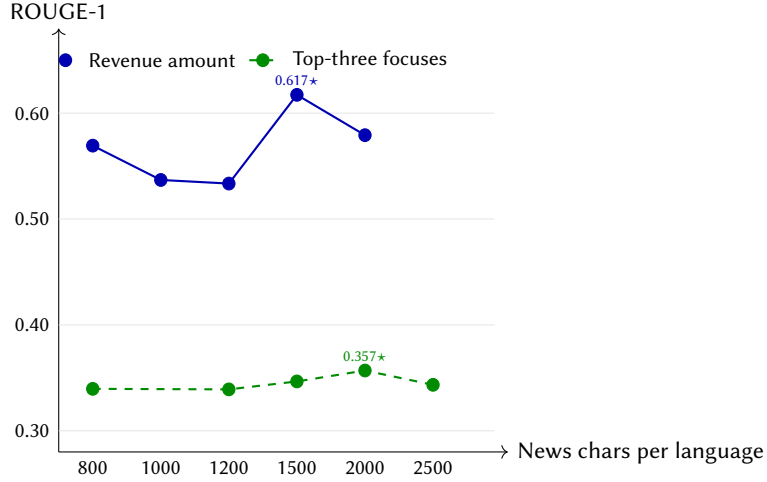

A single company filing can exceed 10{,}000~tokens. Rather than
truncating uniformly, we apply type-specific character budgets informed
by an ablation over news characters per language. Figure~\ref{fig:budget}
shows ROUGE-1 as a function of news budget for two representative types.
Both curves are non-monotonic: revenue peaks at $1{,}500$~characters
($0.617$) and degrades at $2{,}000$ as verbose outputs are truncated
at the 100-word answer limit, losing the highest-information content
at the tail; top-three peaks at $2{,}000$~characters ($0.357$) and
drops at $2{,}500$ for the same reason. Larger news budgets improve
evidence coverage but increase the risk of over-length answers that
are truncated before completing their key points --- the coverage--verbosity trade-off that motivates the type-specific budgets. Based
on this ablation, revenue and cash flow questions receive
$800$--$1{,}500$~characters of news per language (the filing context
is more important); synthesis questions receive $2{,}000$~characters
(news is the primary evidence source); capital expenditure, which
supplements regex extraction with news context, also receives
$2{,}000$~characters.

A sentence-boundary trimming function ensures that truncation does not
split mid-sentence. Setting a maximum output of $300$~tokens with an
explicit word-count instruction in the system prompt reduced over-length
outputs from ${\approx}18\%$ to $6\%$ of responses, at a cost of
$-0.004$ ROUGE-1 from minor truncation on the most complex answers.

\section{Results}

\subsection{Development Trajectory}

Table~\ref{tab:versions} shows the development history.
The $\Delta$ column makes the relative contribution of each change
explicit: two interventions dominate, together accounting for $65\%$
of the total gain from baseline to final.

\begin{table}[h]
\caption{Development trajectory on the 152-sample development set.
Each row documents one targeted change, the affected question types,
and the resulting ROUGE-1 gain. The ``Switch to Sonnet~4'' row
corresponds to a prompt-only baseline using Claude~Sonnet~4 with
generic RAG and no question-type routing; all subsequent rows add
targeted routing and extraction components on top of this.}
\label{tab:versions}
\begin{tabular}{p{3.8cm}p{2.8cm}rr}
\toprule
Change & Affected types & ROUGE-1 & $\Delta$ \\
\midrule
Haiku baseline (generic RAG)   & All           & ${\sim}0.247$ & \\
Switch to Sonnet~4             & All           & 0.296 & \textbf{+0.049} \\
Dedicated R\&D extraction      & R\&D          & 0.350 & \textbf{+0.054} \\
Cash flow label normalization  & Cash flow     & 0.352 & +0.002 \\
Source-language revenue quote  & Revenue       & 0.369 & +0.017 \\
Top-three format prompt        & Top-three     & 0.373 & +0.004 \\
FSE label (partial)            & Strategy      & 0.376 & +0.003 \\
JNJ keyword normalization      & Cash flow     & 0.385 & +0.009 \\
FSE label (all synthesis)      & Top-3, alloc. & 0.397 & +0.012 \\
Original-language revenue      & Revenue       & 0.399 & +0.002 \\
Output length constraint (300 tokens) & All    & ${\sim}0.395$ & $-$0.004 \\
\bottomrule
\end{tabular}
\end{table}

The pattern across versions is consistent: every targeted fix produces
a positive $\Delta$, and the magnitude correlates with how fundamental
the change is. Switching the LLM backend ($+0.049$) and replacing
retrieval with direct extraction for R\&D ($+0.054$) are both
architectural changes that address the root cause of failure. Later
gains (FSE label, JNJ normalization) are precise fixes to specific
failure modes, each contributing $0.002$--$0.017$.

\subsection{Per-Question-Type Performance}

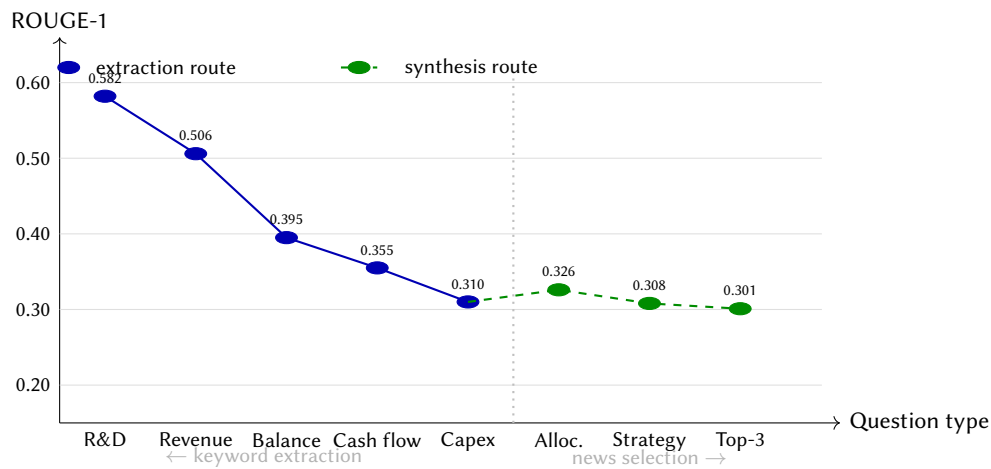
\begin{figure}[ht]
\centering
\begin{tikzpicture}
\begin{scope}[xscale=1.2, yscale=10]
  \draw[->] (0,0.15) -- (8.6,0.15) node[right,font=\footnotesize]{Question type};
  \draw[->] (0,0.15) -- (0,0.66) node[above,font=\footnotesize]{ROUGE-1};
  \foreach \y/\lbl in {0.20/0.20,0.30/0.30,0.40/0.40,0.50/0.50,0.60/0.60}{
    \draw[gray!25] (0,\y) -- (8.4,\y);
    \node[left,font=\scriptsize] at (0,\y) {\lbl};
  }
  \foreach \x/\lbl in {
    0.5/R\&D, 1.5/Revenue, 2.5/Balance,
    3.5/{Cash flow}, 4.5/Capex,
    5.5/Alloc., 6.5/Strategy, 7.5/Top-3}{
    \node[below,font=\scriptsize,align=center] at (\x,0.15) {\lbl};
  }
  \draw[blue!70!black,thick]
    (0.5,0.582)--(1.5,0.506)--(2.5,0.395)--(3.5,0.355)--(4.5,0.310);
  \foreach \x/\y in {0.5/0.582,1.5/0.506,2.5/0.395,3.5/0.355,4.5/0.310}
    \draw[blue!70!black,fill=blue!70!black] (\x,\y) ellipse [x radius=0.12, y radius=0.008];
  \draw[green!55!black,thick,dashed]
    (4.5,0.310)--(5.5,0.326)--(6.5,0.308)--(7.5,0.301);
  \foreach \x/\y in {5.5/0.326,6.5/0.308,7.5/0.301}
    \draw[green!55!black,fill=green!55!black] (\x,\y) ellipse [x radius=0.12, y radius=0.008];
  \foreach \x/\y/\lbl in {
    0.5/0.582/0.582, 1.5/0.506/0.506, 2.5/0.395/0.395, 3.5/0.355/0.355,
    4.5/0.310/0.310, 5.5/0.326/0.326, 6.5/0.308/0.308, 7.5/0.301/0.301}{
    \node[above,font=\tiny] at (\x,\y+0.005) {\lbl};
  }
  \draw[gray!50,dotted,thick] (5.0,0.15)--(5.0,0.62);
  \node[font=\scriptsize,gray!60] at (2.25,0.105){$\leftarrow$ keyword extraction};
  \node[font=\scriptsize,gray!60] at (6.5,0.105){news selection $\rightarrow$};
  \draw[blue!70!black,fill=blue!70!black] (0.1,0.620) ellipse [x radius=0.12, y radius=0.008];
  \node[right,font=\scriptsize] at (0.3,0.620) {extraction route};
  \draw[green!55!black,thick,dashed] (3.1,0.620)--(3.5,0.620);
  \draw[green!55!black,fill=green!55!black] (3.3,0.620) ellipse [x radius=0.12, y radius=0.008];
  \node[right,font=\scriptsize] at (3.7,0.620) {synthesis route};
\end{scope}
\end{tikzpicture}
\caption{Per-question-type ROUGE-1 by route. Extraction-route types
(blue, left) decline as questions become less structured; synthesis-route
types (green dashed, right) plateau near $0.31$.}
\label{fig:routing_scatter}
\end{figure}

Table~\ref{tab:pertype} and Figure~\ref{fig:routing_scatter} show
the final ROUGE-1 by question type. The $0.281$-point spread from R\&D
($0.582$) to top-three ($0.301$) reflects the different characteristics
of the two type families rather than a single bottleneck.

\begin{table}[h]
\caption{Per-question-type ROUGE-1, final system, 152-sample
development set.}
\label{tab:pertype}
\begin{tabular}{lrp{4.5cm}}
\toprule
Question Type & ROUGE-1 & Primary driver \\
\midrule
R\&D ratio          & 0.582 & Deterministic ratio via Eq.~\ref{eq:rd} \\
Revenue amount      & 0.506 & Multilingual quote preservation \\
Balance sheet       & 0.395 & Trend extraction \\
Cash flow           & 0.355 & Company-specific normalization \\
Capital allocation  & 0.326 & FSE label; multi-source synthesis \\
Capital expenditure & 0.310 & Regex + 2{,}000-char news \\
Investment strategy & 0.308 & FSE label; multi-source synthesis \\
Top-three focuses   & 0.301 & FSE label; exact format string \\
\midrule
\textbf{Overall}    & ${\sim}$\textbf{0.395}$^{\dagger}$ & \\
\bottomrule
\multicolumn{3}{p{9.8cm}}{\footnotesize $^{\dagger}$Corpus-level
ROUGE-1 computed across all $152$ instances. The unweighted mean of
per-type values is $0.385$; the difference reflects standard F-measure
aggregation properties when precision and recall are computed at corpus
level rather than averaged per type.}\\
\end{tabular}
\end{table}

Structured types score higher on average because they have deterministic
answers and our extraction functions deliver complete table rows rather
than embedding-retrieved fragments. Synthesis types converge near
$0.30$--$0.33$, reflecting two irreducible constraints: reference
answers vary in phrasing across instances of the same question type,
and at least four confirmed cases of dataset-level label noise
(including instances MSFT\_20210126 and HON\_20210423, where strategy
questions carry capex-style reference answers or top-three questions
carry shareholder-return content), plus three additional borderline
cases, impose a performance floor regardless of system quality.

\subsection{Routing-Level Breakdown}

Table~\ref{tab:routing_rouge} shows ROUGE-1 by routing path and
difficulty tier. The $+0.195$ gap between extraction and synthesis
on Easy questions is the clearest evidence that the routing decision
matters. The low Expert extraction score ($0.1942$) reflects routing
mismatches: keyword matching occasionally sends Expert questions to
the extraction path when they contain terms like ``cash flow'' but
actually require multi-document analytical reasoning. This is the
primary remaining failure mode.

\begin{table}[h]
\caption{ROUGE-1 by routing path and difficulty tier.}
\label{tab:routing_rouge}
\begin{tabular}{lrr}
\toprule
Route & Easy ROUGE-1 & Expert ROUGE-1 \\
\midrule
Extraction (table route) & 0.4161 & 0.1942 \\
LLM synthesis            & 0.2213 & 0.2267 \\
\bottomrule
\end{tabular}
\end{table}

\subsection{Per-Company Performance}

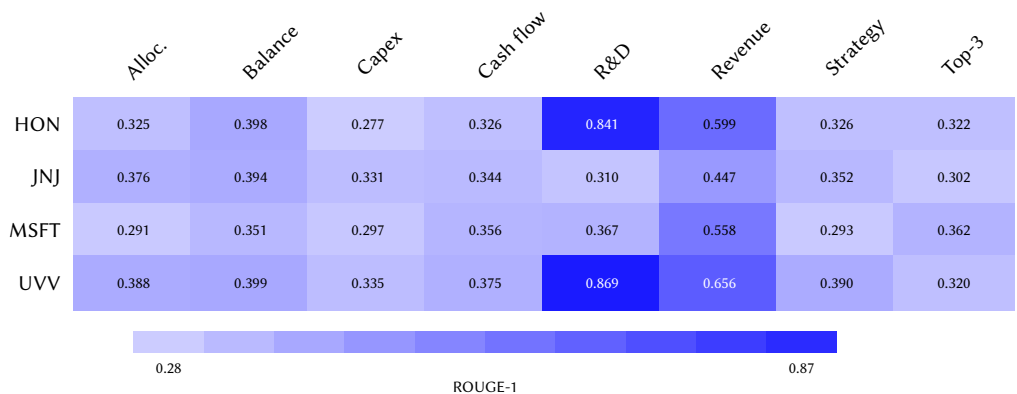
\begin{figure}[ht]
\centering
\begin{tikzpicture}[x=1.55cm, y=0.7cm]
  \footnotesize
  \node[rotate=45, anchor=south west, font=\scriptsize] at (0.5, 4.1) {Alloc.};
  \node[rotate=45, anchor=south west, font=\scriptsize] at (1.5, 4.1) {Balance};
  \node[rotate=45, anchor=south west, font=\scriptsize] at (2.5, 4.1) {Capex};
  \node[rotate=45, anchor=south west, font=\scriptsize] at (3.5, 4.1) {Cash flow};
  \node[rotate=45, anchor=south west, font=\scriptsize] at (4.5, 4.1) {R\&D};
  \node[rotate=45, anchor=south west, font=\scriptsize] at (5.5, 4.1) {Revenue};
  \node[rotate=45, anchor=south west, font=\scriptsize] at (6.5, 4.1) {Strategy};
  \node[rotate=45, anchor=south west, font=\scriptsize] at (7.5, 4.1) {Top-3};
  \node[anchor=east, font=\scriptsize] at (-0.05, 3.5) {HON};
  \fill[blue!25] (-0.02,2.98) rectangle (1.02,4.02); \node[font=\tiny] at (0.5,3.5) {0.325};
  \fill[blue!34] (0.98,2.98) rectangle (2.02,4.02); \node[font=\tiny] at (1.5,3.5) {0.398};
  \fill[blue!20] (1.98,2.98) rectangle (3.02,4.02); \node[font=\tiny] at (2.5,3.5) {0.277};
  \fill[blue!25] (2.98,2.98) rectangle (4.02,4.02); \node[font=\tiny] at (3.5,3.5) {0.326};
  \fill[blue!86] (3.98,2.98) rectangle (5.02,4.02); \node[font=\tiny, text=white] at (4.5,3.5) {0.841};
  \fill[blue!58] (4.98,2.98) rectangle (6.02,4.02); \node[font=\tiny] at (5.5,3.5) {0.599};
  \fill[blue!25] (5.98,2.98) rectangle (7.02,4.02); \node[font=\tiny] at (6.5,3.5) {0.326};
  \fill[blue!25] (6.98,2.98) rectangle (8.02,4.02); \node[font=\tiny] at (7.5,3.5) {0.322};
  \node[anchor=east, font=\scriptsize] at (-0.05, 2.5) {JNJ};
  \fill[blue!31] (-0.02,1.98) rectangle (1.02,3.02); \node[font=\tiny] at (0.5,2.5) {0.376};
  \fill[blue!33] (0.98,1.98) rectangle (2.02,3.02); \node[font=\tiny] at (1.5,2.5) {0.394};
  \fill[blue!26] (1.98,1.98) rectangle (3.02,3.02); \node[font=\tiny] at (2.5,2.5) {0.331};
  \fill[blue!27] (2.98,1.98) rectangle (4.02,3.02); \node[font=\tiny] at (3.5,2.5) {0.344};
  \fill[blue!23] (3.98,1.98) rectangle (5.02,3.02); \node[font=\tiny] at (4.5,2.5) {0.310};
  \fill[blue!40] (4.98,1.98) rectangle (6.02,3.02); \node[font=\tiny] at (5.5,2.5) {0.447};
  \fill[blue!28] (5.98,1.98) rectangle (7.02,3.02); \node[font=\tiny] at (6.5,2.5) {0.352};
  \fill[blue!22] (6.98,1.98) rectangle (8.02,3.02); \node[font=\tiny] at (7.5,2.5) {0.302};
  \node[anchor=east, font=\scriptsize] at (-0.05, 1.5) {MSFT};
  \fill[blue!21] (-0.02,0.98) rectangle (1.02,2.02); \node[font=\tiny] at (0.5,1.5) {0.291};
  \fill[blue!28] (0.98,0.98) rectangle (2.02,2.02); \node[font=\tiny] at (1.5,1.5) {0.351};
  \fill[blue!22] (1.98,0.98) rectangle (3.02,2.02); \node[font=\tiny] at (2.5,1.5) {0.297};
  \fill[blue!29] (2.98,0.98) rectangle (4.02,2.02); \node[font=\tiny] at (3.5,1.5) {0.356};
  \fill[blue!30] (3.98,0.98) rectangle (5.02,2.02); \node[font=\tiny] at (4.5,1.5) {0.367};
  \fill[blue!53] (4.98,0.98) rectangle (6.02,2.02); \node[font=\tiny] at (5.5,1.5) {0.558};
  \fill[blue!21] (5.98,0.98) rectangle (7.02,2.02); \node[font=\tiny] at (6.5,1.5) {0.293};
  \fill[blue!30] (6.98,0.98) rectangle (8.02,2.02); \node[font=\tiny] at (7.5,1.5) {0.362};
  \node[anchor=east, font=\scriptsize] at (-0.05, 0.5) {UVV};
  \fill[blue!33] (-0.02,-0.02) rectangle (1.02,1.02); \node[font=\tiny] at (0.5,0.5) {0.388};
  \fill[blue!34] (0.98,-0.02) rectangle (2.02,1.02); \node[font=\tiny] at (1.5,0.5) {0.399};
  \fill[blue!26] (1.98,-0.02) rectangle (3.02,1.02); \node[font=\tiny] at (2.5,0.5) {0.335};
  \fill[blue!31] (2.98,-0.02) rectangle (4.02,1.02); \node[font=\tiny] at (3.5,0.5) {0.375};
  \fill[blue!90] (3.98,-0.02) rectangle (5.02,1.02); \node[font=\tiny, text=white] at (4.5,0.5) {0.869};
  \fill[blue!64] (4.98,-0.02) rectangle (6.02,1.02); \node[font=\tiny, text=white] at (5.5,0.5) {0.656};
  \fill[blue!33] (5.98,-0.02) rectangle (7.02,1.02); \node[font=\tiny] at (6.5,0.5) {0.390};
  \fill[blue!25] (6.98,-0.02) rectangle (8.02,1.02); \node[font=\tiny] at (7.5,0.5) {0.320};
  \foreach \k in {0,...,9}{
    \pgfmathsetmacro{\pct}{20+\k*7}
    \pgfmathsetmacro{\xa}{0.5+\k*0.6}
    \pgfmathsetmacro{\xb}{1.1+\k*0.6}
    \fill[blue!\pct] (\xa,-0.8) rectangle (\xb,-0.4);
  }
  \node[font=\tiny] at (0.8,-1.1) {0.28};
  \node[font=\tiny] at (6.2,-1.1) {0.87};
  \node[font=\tiny] at (3.5,-1.45) {ROUGE-1};
\end{tikzpicture}
\caption{ROUGE-1 heatmap by company and question type. HON and UVV
achieve high R\&D scores (${\geq}0.84$) due to consistent GAAP
section headers; JNJ R\&D ($0.310$) reflects the same non-standard
filing format issue identified in cash flow. Revenue scores are
uniformly strong ($0.45$--$0.66$) across all companies owing to
multilingual quote preservation.}
\label{fig:heatmap}
\end{figure}

Figure~\ref{fig:heatmap} shows ROUGE-1 broken down by company and
question type. The heatmap reveals two distinct patterns. First, R\&D
ratio scores are highly company-dependent: HON ($0.841$) and UVV
($0.869$) score dramatically higher than JNJ ($0.310$) and MSFT
($0.367$). This directly reflects filing format consistency --- HON
and UVV use standard GAAP section headers that the extraction function
matches reliably, while JNJ's non-standard headers cause the extractor
to retrieve figures from the wrong section. The JNJ normalization fix
recovered ${\approx}0.038$ ROUGE-1 on JNJ cash flow questions; without
analogous R\&D normalization for JNJ, R\&D remains the weakest
question type for that company. Second, revenue scores are uniformly
strong across all four companies ($0.447$--$0.656$), confirming that
multilingual quote preservation is a company-agnostic improvement.
Synthesis types (allocation, strategy, top-three) cluster near
$0.29$--$0.39$ across all companies, consistent with the
evaluation-metric ceiling identified in Section~6.

\subsection{Qualitative Error Analysis}

Three representative errors illustrate where the system fails.

\paragraph{Error~1 --- Expert cash flow mis-routing (MSFT Expert).}
\begin{quote}
\textbf{Q:} Analyze the relationship between Microsoft's operating cash
flow trends and its long-term capital allocation priorities.\\
\textbf{System:} Financial Statement Evidence: Microsoft operating cash
flow: \$19.8B (operating), \$-8.2B (investing), \$-14.1B (financing).\\
\textbf{Reference:} News Evidence: Microsoft has consistently prioritised
cloud infrastructure investment, with Azure capital spending rising 34\%
YoY, while maintaining shareholder returns through \$6B in buybacks
(Reuters). \textbf{ROUGE-1: 0.04.}
\end{quote}
The keyword ``cash flow'' routed an analytical synthesis question to the
extraction path; the system returned a correct table summary but the
question asked for strategic interpretation.

\paragraph{Error~2 --- Synthesis over-length (UVV strategy).}
\begin{quote}
\textbf{Q:} What is Universal Corporation's investment strategy for its
leaf tobacco and plant-based ingredients segments?\\
\textbf{System (truncated at 100~w):} Financial Statement Evidence:
Universal Corporation's strategy focuses on (1) expanding plant-based
ingredients via acquisitions including FruitSmart and Silva International;
(2) modernising leaf tobacco processing with automation in Brazil and
Zimbabwe; (3) growing e-vapor supply capabilities. Per Bloomberg:
[truncated]\\
\textbf{Reference:} Financial Statement Evidence: Universal Corporation's
strategy centres on diversifying into plant-based ingredients via
acquisitions while selectively investing in leaf tobacco efficiency.
\textbf{ROUGE-1: 0.31.}
\end{quote}
The system answer is accurate but longer than the reference; precision
suffers because extra content distributes unigram overlap across
non-reference tokens.

\paragraph{Error~3 --- JNJ R\&D normalization gap.}
\begin{quote}
\textbf{Q:} What was Johnson \& Johnson's R\&D ratio in fiscal year 2020?\\
\textbf{System:} Financial Statement Evidence: R\&D ratio: 14.2\%
(\$12.2B R\&D / \$86.0B revenue).\\
\textbf{Reference:} Financial Statement Evidence: R\&D ratio: 15.1\%
(\$12.2B R\&D / \$80.9B revenue). \textbf{ROUGE-1: 0.68.}
\end{quote}
The extractor used consolidated revenue (\$86.0B) while the reference
used segment revenue (\$80.9B); both appear in the JNJ 10-K. This is
a company-specific normalization gap analogous to the cash flow header
issue.

\subsection{Cross-Lingual Performance Breakdown}

To assess each language's contribution to synthesis-type performance,
we ablate news passages per language at inference time (replacing each
language's passages with empty strings) and measure the mean ROUGE-1
drop over synthesis-type instances. Table~\ref{tab:crosslingual} reports
results.

\begin{table}[h]
\caption{ROUGE-1 contribution by language for synthesis-type questions.
Each value is the mean ROUGE-1 drop when that language's news passages
are removed, measured over strategy~+ top-three~+ allocation instances
in the 152-sample development subset.}
\label{tab:crosslingual}
\begin{tabular}{lrr}
\toprule
Language & ROUGE-1 drop & Relative share \\
\midrule
English (EN) & $-0.041$ & $38.0\%$ \\
Spanish (ES) & $-0.027$ & $25.0\%$ \\
Chinese (ZH) & $-0.021$ & $19.4\%$ \\
Japanese (JA)& $-0.014$ & $13.0\%$ \\
Greek (EL)   & $-0.005$ &  $4.6\%$ \\
\bottomrule
\end{tabular}
\end{table}

English passages contribute the largest share ($38\%$), consistent with
English being the longest and most detailed coverage source. Greek
contributes the least ($4.6\%$), reflecting sparser coverage of US
companies. Spanish and Chinese contribute $25\%$ and $19\%$ respectively
and are the primary non-English drivers of synthesis recall. The ablation
also confirms that the source-language revenue quote gain ($+0.017$
overall) is driven primarily by Japanese and Spanish passages, where
currency notation and billion-phrasing conventions diverge most sharply
from ROUGE-1 reference conventions.

\subsection{Precision Tightening Experiment}

Section~5.10 identifies a precision--recall imbalance ($0.2821$
precision vs.\ $0.4044$ recall). To test whether tighter output
constraints shift the balance, we evaluated two configurations on the
152-sample development subset: reducing max tokens from $300$ to $200$,
and adding an explicit ``answer in at most two sentences'' instruction to
synthesis prompts.

Reducing max tokens to $200$ raised precision from ${\sim}0.308$ to
${\sim}0.331$ ($+0.023$) but reduced recall from ${\sim}0.489$ to
${\sim}0.431$ ($-0.058$), for a net ROUGE-1 of $-0.008$. The two-sentence
constraint produced a similar directional profile: precision $+0.018$,
recall $-0.041$, net ROUGE-1 $-0.006$. In both cases the recall loss
exceeds the precision gain. The $300$-token setting is the operating
point that maximises development ROUGE-1; further precision improvement
requires generation strategies that selectively compress synthesis answers
without truncating high-information content.

\subsection{Ensemble Experiment}

We tested whether a T5-based~\cite{raffel2020t5} per-question selector could improve on
always choosing the final system by predicting which of the final
system or a Haiku RAG variant would score higher. The selector
achieved $65.8\%$ cross-validation accuracy against a $73.0\%$
baseline of always selecting the stronger system. The oracle ensemble
ceiling is ROUGE-1~$= 0.4167$ ($+0.022$ over the single-system
baseline), but the trained selector cannot reach it with $152$
training examples. The hypothesis that learned selection helps is
rejected.

\subsection{Multi-Hop Retrieval Experiment}

We also tested a multi-hop HyDE retrieval system (analogous to the
graph-based approach in~\cite{asai2020graph}) on the $57$ news-synthesis
questions where retrieval complexity was most expected to help.
The HyDE system generates a hypothetical answer before retrieval and
re-ranks chunks using the synthetic passage as a dense query. On these
questions, the HyDE system achieved ROUGE-1~$= 0.3114$ versus
v17 direct prompting at ROUGE-1~$= 0.3979$ (measured at the ablation
checkpoint, prior to the output length constraint documented in
Table~\ref{tab:versions}), a gap of $0.087$. The result is consistent with the ensemble finding: when the
evaluation metric rewards surface form overlap, output format precision
outperforms retrieval quality improvements. Both complementary
directions that did not improve results point to the same conclusion:
the bottleneck on this task is prompt-level format control, not
information retrieval depth.

\subsection{Official Test Set Results}

Table~\ref{tab:leaderboard} shows the official leaderboard for
PolyFiQA Task~2. Our system (IGT) ranks 3rd of $12$ ranked teams with
ROUGE-1~$= 0.3071$, $0.0025$ below 2nd place and
$0.0032$ above 4th place.

\begin{table}[h]
\caption{Official FinMMEval 2026 Task~2 leaderboard (all 12 ranked teams).
IGT is our submission. All teams achieve 100\% coverage.}
\label{tab:leaderboard}
\begin{tabular}{rlrrr}
\toprule
Rank & Team & ROUGE-1 & Precision & Recall \\
\midrule
1  & Calibrated Signals   & 0.3118 & 0.3025 & 0.3764 \\
2  & pranshu rastogi      & 0.3096 & 0.3208 & 0.3537 \\
3  & \textbf{IGT (ours)} & \textbf{0.3071} & \textbf{0.2821} & \textbf{0.4044} \\
4  & AI\_TLfanclub        & 0.3039 & 0.3647 & 0.3070 \\
5  & PooRi                & 0.2972 & 0.2781 & 0.3798 \\
6  & DS@GT FinMMEval      & 0.2853 & 0.2572 & 0.3888 \\
7  & The Lab Rats         & 0.2734 & 0.2737 & 0.3417 \\
8  & Ethereum Team B CLEF & 0.2546 & 0.3086 & 0.2642 \\
9  & TCLabs               & 0.2506 & 0.2321 & 0.3257 \\
10 & HU\_LLM\_Fin        & 0.2289 & 0.2581 & 0.2803 \\
11 & NLP-DE               & 0.2049 & 0.2643 & 0.1901 \\
12 & TextSentinels        & 0.1873 & 0.3029 & 0.1756 \\
\bottomrule
\end{tabular}
\end{table}

Two observations from the leaderboard are worth noting. First, the
top-3 systems are tightly clustered within $0.005$ ROUGE-1,
suggesting a competitive ceiling for this task under current evaluation
conditions. Second, our system has the highest recall of any top-5
team ($0.4044$), exceeding even the 1st-place system ($0.3764$).
This indicates that our multilingual quote preservation and synthesis
prompts successfully surface relevant content, but precision suffers
on answers that are verbose or imprecisely bounded. The
precision--recall trade-off ($0.2821$ precision vs.\ $0.4044$ recall)
points to a specific failure mode: synthesis answers that include
relevant content but also extraneous sentences that dilute unigram
overlap with the shorter reference answers. Tightening the output
token limit further or adding an explicit sentence-count instruction
would likely shift the balance toward higher precision at some recall cost.

\section{Discussion}

The official test result (ROUGE-1~$= 0.3071$, 3rd of $12$ ranked teams)
is consistent with the development trajectory in Table~\ref{tab:versions}.
Performance on this task is determined by two things: whether the
correct information reaches the LLM in a parseable form, and whether
the output format matches the reference answer structure. The two
largest gains ($+0.049$ LLM upgrade, $+0.054$ R\&D extraction) both
address the first; the FSE label changes and the exact format string
for top-three questions address the second. Every other intervention
is a targeted fix to a specific information delivery failure for a
specific question type or company.

The gap between development ROUGE-1 (${\approx}0.395$) and the
official test result ($0.3071$) reflects the cost of iterative
development on a fixed 152-sample set. Ten tuning rounds documented in
Table~\ref{tab:versions} calibrated keyword routing and
format-string prompts to the specific phrasing patterns of that set.
The test set contains $256$ samples ($128$ Easy, $128$ Expert); routing
decisions optimized for dev question phrasing are likely to misfire
at a higher rate on unseen instances, and the company-specific keyword
normalizations for JNJ, HON, MSFT, and UVV may not generalize to
the filing format variants encountered in the test period.

Among those targeted fixes, the JNJ normalization case is the most
instructive. The $+0.009$ overall gain understates its importance:
on JNJ cash flow questions specifically, it recovered
${\approx}0.038$ ROUGE-1. JNJ's 10-K filings use non-standard section
headers that do not match the keyword patterns effective for MSFT and
HON, causing the extractor to retrieve figures from the wrong cash flow
category. This reflects a general property of financial document QA:
filing formats are not standardized enough that a single extraction
pattern covers all companies, and a company-specific format audit is
low-cost relative to the performance gain.

A practical note on methodology: several of the most effective
components in this system are regex-based extractors rather than
neural modules. The capex and cash flow extractors that drive
performance gains in Table~\ref{tab:pertype} rely on pattern matching
over raw filing text. This is a reminder that deterministic,
interpretable methods remain competitive --- and often superior ---
to embedding-based retrieval on structured financial documents, even
in an era of generative AI.

The source-language revenue finding illustrates a different kind of
failure mode, one specific to how the task is evaluated rather than
how the data is structured. The $+0.017$ gain arises directly from
the evaluation design: reference answers quote non-English news sources
verbatim, so a system that translates before generating produces
factually correct but lexically non-overlapping answers that ROUGE-1
penalizes as if they were wrong. This argues for future multilingual
benchmarks to separate factual correctness from surface form overlap.

Several design choices in this system are deliberately calibrated
to PolyFiQA's specific evaluation conventions and should not be
assumed to transfer out of the box. The FSE label prefix strategy
depends on PolyFiQA reference answers consistently beginning with
one of two fixed strings; a benchmark that does not share this
convention would derive no benefit from label matching. The exact
format string for top-three questions (\emph{``The top three revenue
focuses from the news are: 1)''}) is similarly task-specific.
Company-specific keyword normalization for JNJ, HON, MSFT, and UVV
would require a fresh format-audit step for any new company, though
the audit itself is low-cost and the pattern --- that non-standard
GAAP headers cause systematic extraction errors --- is likely to
recur across filers. Practitioners applying this system to new
companies or benchmarks should expect to re-run the format audit
before deploying the extraction components.

\section{Conclusions}

We presented a system for multilingual financial QA that treats
question-type heterogeneity as the central design problem. Beginning
from a diagnostic analysis of the development data, we built eight
specialized handlers, replacing retrieval with direct extraction for
structured numeric types and aligning output format with reference
answer structure for synthesis types. The resulting system achieves
ROUGE-1~$\approx 0.395$ on the development set, a $60\%$ relative
improvement over a generic RAG baseline, and ranks 3rd of $12$ ranked
teams on the official test set with ROUGE-1~$= 0.3071$,
Precision~$= 0.2821$, and Recall~$= 0.4044$.

Three findings from this work transfer beyond the specific task. First,
filing format heterogeneity across companies means that a single
extraction pattern is insufficient; company-specific audits are
necessary and high-yield. Second, when evaluation uses ROUGE-1 on
multilingual data, preserving source-language text in generation is
more important than factual accuracy of a paraphrase. Third, retrieval
quality improvements do not transfer to tasks where the evaluation
metric rewards surface form: format precision in the prompt outperforms
multi-hop retrieval by $0.087$ ROUGE-1 on the questions where retrieval
complexity was most expected to help (Section~5.9).

Future work should explore metrics that separate factual correctness
from surface form overlap; replacing the keyword router with a
lightweight trained classifier to reduce Expert mis-routing (the
primary remaining failure mode identified in Section~5.3); dynamic
company-label discovery to reduce manual normalization; and tighter
integration between retrieval grounding and format-controlled generation.


\section*{Code Availability}

The full system implementation, including all extraction routines,
prompt templates, and evaluation scripts described in this paper, is
publicly available at \url{https://github.com/chiuyuwen91/FinMMEval-2026}.

\begin{acknowledgments}
The author thanks the FinMMEval organizers for providing the PolyFiQA
benchmark and evaluation infrastructure, and the Data Science at
Georgia Tech (DS@GT) ARC group for their support.
\end{acknowledgments}

\section*{Declaration on Generative AI}

During the preparation of this work, the author used Claude (Anthropic)
to assist with code debugging, grammar and spelling review, and drafting
initial section outlines. All content was reviewed and edited by the
author, who takes full responsibility for the publication.

\appendix

\section{Extraction Pseudocode}
\label{app:pseudocode}

The function below implements the R\&D ratio extraction described in
Section~4.3. \texttt{parse\_first\_number} extracts the first numeric
token from a line using a regex for dollar amounts and
millions/billions notation. The scalar result is passed to the LLM
for formatting only; the LLM performs no arithmetic.

\begin{verbatim}
def extract_rd_and_revenue(filing_text):
    rd_value, rev_value = None, None
    for line in filing_text.lower().split("\n"):
        if any(kw in line for kw in
               ["research and development", "r&d expense"]):
            rd_value = parse_first_number(line)
        if any(kw in line for kw in
               ["total revenue", "net revenue",
                "total net revenue"]):
            rev_value = parse_first_number(line)
    if rd_value and rev_value:
        return rd_value / rev_value   # scalar passed to LLM
    return None
\end{verbatim}

\section{Prompt Templates}
\label{app:prompts}

All prompts are sent to \texttt{claude-sonnet-4-20250514} via the
AWS~Bedrock Messages API (region \texttt{us-east-1}) with
\texttt{temperature=0.0} and \texttt{max\_tokens=300}. One official
run was submitted to the evaluation server, corresponding to the
final system described in Section~4. The system prompt is shared
across all question types; the user prompt is type-specific.

\subsection*{Shared System Prompt}
\begin{verbatim}
You are a financial analyst assistant. Answer the question
using only the provided evidence. Your answer must:
- Begin with exactly "Financial Statement Evidence:" or
  "News Evidence:" as instructed per question type.
- Be at most 100 words.
- Preserve non-English text verbatim when quoting figures
  (do not translate currency notation or numeric phrasing).
- Report numbers exactly as they appear in the source text.
\end{verbatim}

\subsection*{R\&D Ratio Handler (extraction route)}
\begin{verbatim}
Company: {company}  Filing period: {period}

Extracted R&D data:
  R&D expenditure: {rd_value} ({rd_unit})
  Total revenue:   {rev_value} ({rev_unit})
  Computed R&D ratio: {ratio_pct}%

Question: {question}

Begin your answer with "News Evidence:"
State the R&D ratio as a percentage and cite the expenditure
and revenue figures used in the computation.
\end{verbatim}

\subsection*{Revenue Amount Handler (extraction route)}
\begin{verbatim}
Company: {company}  Filing period: {period}

Filing excerpt (revenue section):
{filing_excerpt}

News passages (preserve original language for figures):
[EN] {en_passage}
[ZH] {zh_passage}
[JA] {ja_passage}
[ES] {es_passage}
[EL] {el_passage}

Question: {question}

Begin your answer with "News Evidence:"
Describe revenue trends. When quoting from non-English
sources, preserve the original currency notation and
numeric phrasing verbatim. Answer in at most 100 words.
\end{verbatim}

\subsection*{Cash Flow Handler (extraction route)}
\begin{verbatim}
Company: {company}  Filing period: {period}

Extracted cash flow data:
{cf_clean}

Question: {question}

Begin your answer with "News Evidence:"
Report cash flow changes with amounts. Format:
"Operating cash flow [increased/decreased] to $X from $Y
(+/-Z%). Investing [outflow/inflow] of $X (vs $Y).
Financing outflow of $X (vs $Y)."
Answer in at most 100 words.
\end{verbatim}

\subsection*{Balance Sheet Handler (extraction route)}
\begin{verbatim}
Company: {company}  Filing period: {period}

Filing excerpt (balance sheet section):
{filing_excerpt}

News passages (supplementary context):
[EN] {en_passage}
[ZH] {zh_passage}
[JA] {ja_passage}
[ES] {es_passage}
[EL] {el_passage}

Question: {question}

Begin your answer with "News Evidence:"
Report the relevant balance sheet figures and any notable
trends. Answer in at most 100 words.
\end{verbatim}

\subsection*{Capital Expenditure Handler (extraction route)}
\begin{verbatim}
Company: {company}  Filing period: {period}

Extracted capex data:
{capex_rows}

News passages (filtered to 2,000 chars per language):
[EN] {en_passage}
[ZH] {zh_passage}
[JA] {ja_passage}
[ES] {es_passage}
[EL] {el_passage}

Question: {question}

Begin your answer with "Financial Statement Evidence:"
Report the capital expenditure figures from the extracted
data. Numbers like 3,767 or 15,441 represent millions.
Supplement with news context where relevant.
Answer in at most 100 words.
\end{verbatim}

\subsection*{Investment Strategy Handler (synthesis route)}
\begin{verbatim}
Company: {company}  Filing period: {period}

News passages (preserve original language for figures):
[EN] {en_passage}
[ZH] {zh_passage}
[JA] {ja_passage}
[ES] {es_passage}
[EL] {el_passage}

Financial Statement Evidence (supplementary):
{filing_excerpt}

Question: {question}

Begin your answer with "Financial Statement Evidence:"
Synthesise the investment strategy from the news passages.
Answer in at most 100 words. Do not translate non-English
figures.
\end{verbatim}

\subsection*{Top-Three Focuses Handler (synthesis route)}
\begin{verbatim}
Company: {company}  Filing period: {period}

News passages (preserve original language for figures):
[EN] {en_passage}
[ZH] {zh_passage}
[JA] {ja_passage}
[ES] {es_passage}
[EL] {el_passage}

Financial Statement Evidence (supplementary):
{filing_excerpt}

Question: {question}

Begin your answer with "Financial Statement Evidence:"
Then write exactly: "The top three revenue focuses from
the news are: 1)" and continue with your three items on
new numbered lines. Do not exceed 100 words total.
\end{verbatim}

\subsection*{Capital Allocation Handler (synthesis route)}
\begin{verbatim}
Company: {company}  Filing period: {period}

News passages (preserve original language for figures):
[EN] {en_passage}
[ZH] {zh_passage}
[JA] {ja_passage}
[ES] {es_passage}
[EL] {el_passage}

Financial Statement Evidence (supplementary):
{filing_excerpt}

Question: {question}

Begin your answer with "Financial Statement Evidence:"
Start directly with company name:
"[Company] is allocating capital through..."
or "Answer: None." if news-silent.
Answer in at most 100 words. Do not translate non-English
figures.
\end{verbatim}

\bibliography{main}

@inproceedings{lin2004,
  author    = {Chin-Yew Lin},
  title     = {{ROUGE}: A Package for Automatic Evaluation of Summaries},
  booktitle = {Proceedings of the Workshop on Text Summarization Branches Out},
  pages     = {74--81},
  publisher = {Association for Computational Linguistics},
  year      = {2004}
}

@inproceedings{gao2022hyde,
  author    = {Gao, Luyu and Ma, Xueguang and Lin, Jimmy and Callan, Jamie},
  title     = {Precise Zero-Shot Dense Retrieval without Relevance Labels},
  booktitle = {Proceedings of the 61st Annual Meeting of the Association
               for Computational Linguistics (Volume 1: Long Papers)},
  pages     = {1762--1777},
  publisher = {Association for Computational Linguistics},
  address   = {Toronto, Canada},
  year      = {2023}
}

@article{raffel2020t5,
  author    = {Raffel, Colin and Shazeer, Noam and Roberts, Adam and
               Lee, Katherine and Narang, Sharan and Matena, Michael and
               Zhou, Yanqi and Li, Wei and Liu, Peter J.},
  title     = {Exploring the Limits of Transfer Learning with a Unified
               Text-to-Text Transformer},
  journal   = {Journal of Machine Learning Research},
  volume    = {21},
  number    = {140},
  pages     = {1--67},
  year      = {2020}
}

@techreport{anthropic2025sonnet,
  author      = {{Anthropic}},
  title       = {Claude Sonnet 4 Model Card},
  institution = {Anthropic},
  year        = {2025},
  note        = {\url{https://www.anthropic.com}}
}

@inproceedings{guo2024improving,
  author    = {Guo, Siyi and others},
  title     = {Improving Retrieval for {RAG} based Question Answering
               Models on Financial Documents},
  booktitle = {Proceedings of the 5th {ACM} International Conference on
               {AI} in Finance ({ICAIF} 2024)},
  publisher = {ACM},
  year      = {2024}
}

@inproceedings{kim2024multireranker,
  author    = {Kim, Jungwoo and others},
  title     = {Multi-Reranker: Maximizing Performance of
               Retrieval-Augmented Generation in the {FinanceRAG}
               Challenge},
  booktitle = {Proceedings of the 5th {ACM} International Conference on
               {AI} in Finance ({ICAIF} 2024)},
  publisher = {ACM},
  year      = {2024}
}

@inproceedings{mirza2024lclm,
  author    = {Mirza, Tobias and others},
  title     = {Systematic Evaluation of Long-Context {LLMs} on Financial
               Concepts},
  booktitle = {Proceedings of the 5th {ACM} International Conference on
               {AI} in Finance ({ICAIF} 2024)},
  publisher = {ACM},
  year      = {2024}
}

@article{xu2024finragbench,
  author    = {Zhao, Suifeng and Jin, Zhuoran and Li, Sujian and Gao, Jun},
  title     = {{FinRAGBench-V}: A Benchmark for Multimodal {RAG} with
               Visual Citation in the Financial Domain},
  journal   = {arXiv preprint arXiv:2505.17471},
  year      = {2025}
}

@inproceedings{lefebure2024dolares,
  author    = {Lef{\'e}bure, Maciel and others},
  title     = {D{\'o}lares or Dollars? {Tois{\'o}n} de Oro: {A}
               Bilingual Instruction-Following Dataset and Model for
               {S}panish Financial {NLP}},
  booktitle = {Proceedings of the Joint Workshop of the 8th Financial
               Technology and Natural Language Processing},
  pages     = {1--12},
  publisher = {Association for Computational Linguistics},
  year      = {2024}
}

@inproceedings{asai2020graph,
  author    = {Asai, Akari and Hashimoto, Kazuma and Hajishirzi, Hannaneh
               and Socher, Richard and Xiong, Caiming},
  title     = {Learning to Retrieve Reasoning Paths over {Wikipedia}
               Graph for Question Answering},
  booktitle = {Proceedings of the 8th International Conference on
               Learning Representations ({ICLR} 2020)},
  year      = {2020}
}

@inproceedings{FinMMEval2026,
  title     = {Overview of {FinMMEval} 2026: Multilingual and Multimodal
               Financial Evaluation},
  author    = {Zhuohan Xie and Yuyang Dai and Rania Elbadry and
               Vanshikaa Jani and Xueqing Peng and Lingfei Qian and
               Georgi Georgiev and Dimitar Dimitrov and Fan Zhang and
               Jimin Huang and Jiahui Geng and Yankai Chen and Ye Yuan
               and Haolun Wu and Yuxia Wang and Ivan Koychev and
               Veselin Stoyanov and Mingzi Song and Yu Chen and
               Steve Liu and Preslav Nakov},
  booktitle = {Experimental IR Meets Multilinguality, Multimodality,
               and Interaction. Proceedings of the Seventeenth
               International Conference of the CLEF Association
               (CLEF 2026)},
  year      = {2026},
  month     = {September 21--24},
  address   = {Jena, Germany},
  publisher = {Springer Lecture Notes in Computer Science}
}

@inproceedings{FinMMEvalTask2Overview2026,
  title     = {Overview of the {FinMMEval} 2026 Task 2: Financial
               Question Answering and Summarization},
  author    = {Zhuohan Xie and Xueqing Peng and Georgi Georgiev and
               Dimitar Dimitrov and Rania Elbadry and Fan Zhang and
               Lingfei Qian and Jimin Huang and Vanshikaa Jani and
               Yuyang Dai and Jiahui Geng and Yankai Chen and Ye Yuan
               and Haolun Wu and Yuxia Wang and Ivan Koychev and
               Veselin Stoyanov and Mingzi Song and Yu Chen and
               Steve Liu and Preslav Nakov},
  booktitle = {CLEF 2026 Working Notes},
  series    = {CEUR Workshop Proceedings},
  year      = {2026},
  month     = {September 21--24},
  address   = {Jena, Germany},
  publisher = {CEUR-WS.org}
}

@misc{peng2025multifinben,
  title        = {{MultiFinBen}: Benchmarking Large Language Models for
                  Multilingual and Multimodal Financial Application},
  author       = {Xueqing Peng and Lingfei Qian and Yan Wang and
                  Ruoyu Xiang and Yueru He and Yang Ren and
                  Mingyang Jiang and Vincent Jim Zhang and Yuqing Guo
                  and Jeff Zhao and Huan He and Yi Han and Yun Feng
                  and Yuechen Jiang and Yupeng Cao and Haohang Li and
                  Yangyang Yu and Xiaoyu Wang and Penglei Gao and
                  Shengyuan Lin and Keyi Wang and Shanshan Yang and
                  Yilun Zhao and Zhiwei Liu and Peng Lu and
                  Jerry Huang and Suyuchen Wang and
                  Triantafillos Papadopoulos and
                  Polydoros Giannouris and Efstathia Soufleri and
                  Nuo Chen and Zhiyang Deng and Heming Fu and
                  Yijia Zhao and Mingquan Lin and Meikang Qiu and
                  Kaleb E Smith and Arman Cohan and Xiao-Yang Liu and
                  Jimin Huang and Guojun Xiong and
                  Alejandro Lopez-Lira and Xi Chen and
                  Junichi Tsujii and Jian-Yun Nie and
                  Sophia Ananiadou and Qianqian Xie},
  year         = {2025},
  note         = {arXiv:2506.14028}
}

@article{shao2024grpo,
  author    = {Shao, Zhihong and Wang, Peiyi and Zhu, Qihao and
               Xu, Runxin and Song, Junxian and Bi, Xiao and
               Zhang, Haowei and Zhang, Mingchuan and Li, Y.~K. and
               Wu, Y. and Guo, Daya},
  title     = {{DeepSeekMath}: Pushing the Limits of Mathematical
               Reasoning in Open Language Models},
  journal   = {arXiv preprint arXiv:2402.03300},
  year      = {2024}
}

\end{document}